\documentclass[sigconf]{acmart}
\usepackage{graphicx} 
\usepackage{stfloats}
\usepackage{natbib}
\usepackage{hyperref}
\usepackage{enumitem}

\usepackage{amssymb}
\usepackage{hyperref}

\AtBeginDocument{%
  }

\copyrightyear{2023}
\acmYear{2023}
\setcopyright{acmlicensed}\acmConference[SIGIR-AP '23]{Annual International ACM SIGIR Conference on Research and Development in Information Retrieval in the Asia Pacific Region}{November 26--28, 2023}{Beijing, China}
\acmBooktitle{Annual International ACM SIGIR Conference on Research and Development in Information Retrieval in the Asia Pacific Region (SIGIR-AP '23), November 26--28, 2023, Beijing, China}
\acmPrice{15.00}
\acmDOI{10.1145/3624918.3625338}
\acmISBN{979-8-4007-0408-6/23/11}

\date{May 2023}

\begin{document}

\title[Open-Domain Dialogue Quality Evaluation: Deriving Nugget-level Scores from Turn-level Scores]{Open-Domain Dialogue Quality Evaluation: Deriving Nugget-level Scores from Turn-level Scores}

\author{Rikiya Takehi}
\email{rikiya.takehi@fuji.waseda.jp}
\orcid{0009-0003-6336-8064}
\affiliation{%
  \institution{Waseda University}
  \country{Japan}
}

\author{Akihisa Watanabe}
\email{akihisa@ruri.waseda.jp}
\affiliation{%
  \institution{Waseda University}
  \country{Japan}
}

\author{Tetsuya Sakai}
\email{tetsuyasakai@acm.org}
\affiliation{%
  \institution{Waseda University}
  \country{Japan}
}

\begin{abstract}
Existing dialogue quality evaluation systems can return a score for a given system turn from a particular viewpoint, e.g., engagingness. However, to improve dialogue systems by locating exactly where in a system turn potential problems lie, a more fine-grained evaluation may be necessary. We therefore propose an evaluation approach where a turn is decomposed into nuggets (i.e., expressions associated with a dialogue act), and nugget-level evaluation is enabled by leveraging an existing turn-level evaluation system. We demonstrate the potential effectiveness of our evaluation method through a case study.\footnote{This is a preprint of our upcoming SIGIR-AP 2023 paper. Please cite the SIGIR-AP version.}
\end{abstract}


\maketitle

\section{Introduction}\label{s:intro}
The recent development of Large Language Models has increased the daily use of open-domain dialogue systems like conversational search \cite{10.1145/3020165.3020183} and conversational recommender systems \cite{10.1145/3209978.3210002}. This has led to an increase in the necessity of evaluating these models \cite{see-etal-2019-makes}. However, evaluating these systems is not an easy task, since various dialogue qualities need to be considered \cite{10.1007/s10462-020-09866-x, phy-etal-2020-deconstruct, yeh2021comprehensive}. Moreover, while some dialogue qualities are considered to be conversation-level dialogue qualities, like diversity\cite{ma-etal-2022-self, chen2023exploring} and consistency \cite{zhang-etal-2021-dynaeval}, many qualities like engagingness\cite{10.1145/3576840.3578319} and interestingness \cite{zhang-etal-2020-dialogpt}, are often defined to be turn-level dialogue qualities.

Some recent frameworks \cite{ghazarian2020predictive, yi-etal-2019-towards, 10.1145/3576840.3578319, xu-etal-2022-endex, phy-etal-2020-deconstruct} have been developed to assess turn-level dialogue qualities, effectively evaluating the quality of each turn within a dialogue. While some take a classification approach for this task by categorizing the quality of the turns into predefined classes \cite{ghazarian2020predictive, yi-etal-2019-towards}, many frameworks employ a regression-based approach, where they assign a dialogue quality score to each system turn in the dialogue \cite{10.1145/3576840.3578319, xu-etal-2022-endex, phy-etal-2020-deconstruct}. Additionally, there are regression-based frameworks that are typically used to derive turn-level quality scores for an entire dialogue \cite{ma-etal-2022-self, zhang-etal-2021-dynaeval}, but can also be adapted to give a score of these qualities in each system turn by inputting one interaction at a time. These regression-based turn-level dialogue quality evaluation frameworks, whether they are used to assign scores to individual turns or to the whole dialogue, can achieve precise turn-level evaluations and high scalability \cite{10.1145/3576840.3578319}.

However, as LLMs have started to respond longer turns, the specificity of the evaluation has become more crucial. Analyzing dialogue qualities at the finer specificity, like sentence level and phrase level, would lead to more precise insights into the strengths and weaknesses of a dialogue system, which in turn can be instrumental in improving its practical application \cite{sakai2023swan}. Particularly, while there are several turn-level dialogue quality evaluation frameworks, these frameworks should also be able to spot the problematic sections within the turn to precisely analyze and modify the system when trouble occurs.

To address this need for more fine-grained analysis, our focus in this research is to provide a method to empower the specificity of regression-based turn-level dialogue quality evaluation frameworks. In our approach, we accomplish this by obtaining a nugget-level score from turn-level scores. A nugget may represent a factual statement or a dialogue act~\cite{sakai2023swan}; the present study considers the latter type.

In this research, we first provide a taxonomy to classify sentences in a system turn into corresponding dialogue acts. Then, we propose our method for obtaining a nugget-level score based on a turn-level score. Finally, we provide a simple application example of our method and its results.

The contributions of this paper are three-fold:
\begin{itemize}
    \item To the best of our knowledge, our research is the first to evaluate open-domain dialogue systems to the nugget level. Unlike the previous studies, our approach enables not only turn-level evaluations but also evaluations at a finer level within each turn.
    \item We design and implement a method to derive a nugget-level dialogue quality score from turn-level scores. Our case study shows the ability of our method in giving a more precise insight into the strengths and weaknesses of a dialogue system.
    \item We introduce a new taxonomy for classifying dialogue acts in responses generated by open-domain dialogue systems, which facilitates a better understanding and analysis of these responses.
\end{itemize}

\section{related work}
\subsection{Dialogue Qualities}
Although many dialogue qualities can be considered appropriate to be evaluated at the nugget level for more fine-grained analysis, not all of them are suitable for such assessment \cite{sakai2023swan}. Conversation-level dialogue qualities like diversity \cite{zhang-etal-2021-dynaeval, wu-etal-2020-diverse}, are qualities that are only evaluated at the conversation level and never at a finer level; these qualities are therefore not suitable for nugget-level analysis. Even some turn-level dialogue qualities like conciseness, which is a quality of how minimal the system turn length is, can only be evaluated at the turn level and not at the nugget level. In contrast, some dialogue qualities are more suitable to be evaluated solely at the nugget level and should not be derived based on the analysis of the turn-level score. Specifically, more well-characterized and factual categories that are not related to dialogue acts, like correctness, are able to be evaluated solely at the nugget level because they can be analyzed by only focusing on every single nugget and do not need to consider correlations with other sentences or turns. 

However, there are several qualities such as engagingness \cite{xu-etal-2022-endex, 10.1145/3576840.3578319, zhang-etal-2021-dynaeval}, relevance \cite{mehri-eskenazi-2020-unsupervised, shin2021generating, zhang-etal-2020-dialogpt}, and interestingness \cite{mehri-eskenazi-2020-unsupervised, see-etal-2019-makes, ma-etal-2022-self, zhang-etal-2021-dynaeval} that could benefit from nugget-level scores derived from turn-level evaluations. These qualities are often abstract, multi-dimensional, and may involve correlations with other sentences and turns \cite{see-etal-2019-makes}, making direct nugget-level analysis challenging. Despite the potential benefits of more specific evaluations for these qualities, to the best of our knowledge, existing approaches focus only on the conversation level or turn level, by training on data like human annotations \cite{ghazarian2020predictive, yi-etal-2019-towards}, weak labels \cite{10.1145/3576840.3578319} and human reactions \cite{xu-etal-2022-endex}.

Our proposed method seeks to obtain nugget-level scores by effectively using an existing turn-level evaluation framework, targeting those qualities that are abstract and difficult to evaluate directly at the nugget level.

\subsection{Nugget-Based Evaluation} 
Nugget-based evaluation has been previously proposed and applied in various tasks including search result summarization \cite{10.1145/2063576.2063669, 10.1145/2484028.2484031} and task-oriented dialogue systems \cite{tao2022overview}. For instance, Sakai et al. \cite{10.1145/2063576.2063669} evaluated search result summaries by breaking down the summarized text into nuggets and assigning weights to these nuggets based on their importance and position within the text. However, tasks like search summarization often require each nugget to be evaluated based on a single criterion of `importance'. In contrast, open-domain dialogue systems may require evaluation based on multiple dialogue qualities \cite{sakai2023swan}.

Our research aims to enable the weighting of nuggets based on multiple dialogue qualities, which holds the promise of significantly advancing the application of nugget-based evaluations in open-domain dialogue systems.

\section{Taxonomy for Nugget Classification}
\begin{table}
    \caption{Taxonomy of Dialogue Acts and Corresponding Examples for Nugget in Open-Domain Dialogue Systems}
    \label{tab:labels}
    \begin{tabular}{ccl}
        \toprule
        Dialogue Act & Example\\
        \midrule
        Agreement & I agree \\
        Disagreement & I disagree \\
        Yes Answer & Yes, you are correct \\
        No Answer & No, that is wrong \\
        Opening & Hello \\
        Closing & It was nice talking with you. \\
        Apology & I am sorry \\
        Thanking & Thank you \\
        Rejection & I cannot provide an answer. \\
        Applause & Well done. \\
        Declarative Question & What do you mean by ...?\\
        Confusion & I don't understand \\
        Reasoning & This is because ... \\
        Downplayer & That's all right. \\
        Assumption & I assume you meant ... \\
        Acknowledgment & Ok. \\
        Clarification & The pdf you provided me is .... \\
        Non-Declarative Question & Isn't it exciting? \\
        User instruction & Please click on .... \\
        Recommendation & I would recommend.... \\
        Citation & According to ... \\
        Example & For example, ... \\
        Commissive & I am happy to help ... \\
        Opinion & I think ...\\
        \bottomrule
    \end{tabular}
\end{table}

In order to perform a fine-grained evaluation of open-domain dialogue system responses, it is imperative to analyze sentences at a more granular level. To this end, we introduce a taxonomy that allows us to classify each sentence into labeled nuggets. A nugget in this context refers to a segment of a sentence that embodies a single dialogue act. Dialogue acts represent the intention or illocutionary force behind an utterance. It is important to note that a single sentence may encompass multiple dialogue acts, leading to the extraction of several nuggets. For instance, consider a sentence ``I am sorry, I cannot provide an answer for that.'' This sentence can be decomposed into two nuggets, namely an \textbf{Apology} nugget ``I am sorry,'' and a \textbf{Rejection} nugget ``I cannot provide an answer for that'' \cite{sakai2023swan}.

Our taxonomy, which is specifically tailored for classifying sentences in open-domain dialogue system responses, is based on the taxonomy of dialogue acts introduced by Stolcke et al. \cite{stolcke-etal-2000-dialogue} for conversational speech classification. However, given the distinct nature of open-domain dialogue systems, we have adapted the original taxonomy by excluding non-verbal dialogue acts, such as laughter, and focusing on dialogue acts that are more pertinent to textual dialogue systems. This results in a taxonomy that is more focused and relevant to the evaluation of dialogue system responses.

Table~\ref{tab:labels} presents the taxonomy of dialogue acts for open-domain dialogue system response nugget classification. Each dialogue act is accompanied by an example.

This taxonomy is designed to serve as a robust tool for evaluating and understanding open-domain dialogue system responses. 

\section{Our approach}
Our approach necessitates the use of an existing turn-level evaluation framework (eg. \cite{xu-etal-2022-endex, mehri-eskenazi-2020-unsupervised, 10.1145/3576840.3578319}), that can give a score of a given interaction. Using these turn-level evaluation frameworks, we will evaluate each nugget by analyzing how much each nugget is contributing to the turn-level evaluation, which is a direct indication of the nugget-level score. The nugget-level evaluation score will be calculated in the following procedures.

We treat a system turn as a set of nuggets $n$, which we denote as $T=\{n, ...\}$. Then, given a system turn $T$ and a turn-level evaluation framework, the most demonstrative approach to estimate an evaluation score of each nugget $n (\in T)$ would be to delete the nugget and see how the turn-level evaluation score has changed from the original turn-level score $s(T)$. We denote Turn $T$ with nugget $n(\in T)$ deleted as $T_{n\to\phi}$. Then, we use $s(T_{n\to\phi})$ to denote the turn-level evaluation score when deleting nugget $n(\in T)$. In simple cases, if $s(T_{n\to\phi})$ turns out to be smaller than $s(T)$, it is probable that nugget $n(\in T)$ is positively influencing the turn-level score; therefore, this nugget $n(\in T)$ should be given a high nugget-level score. We define this difference of scores, $D_{\phi}(T,n)$, as:
\begin{equation}
    D_{\phi}(T,n)=s(T)-s(T_{n\to\phi}) .
    \label{eq:MSemp}
\end{equation}

Deletion of the nugget alone, however, would not be enough to give a precise evaluation score, because there could be more appropriate nuggets of other dialogue acts than the original nugget $n(\in T)$. If there are alternative nuggets of different dialogue acts that make a better turn-level score, the score of nugget $n(\in T)$ should decrease. Turn $T$ with nugget $n(\in T)$ replaced with a generated nugget $n_{\mathrm{diff}}$ whose dialogue act is \textit{different} from that of $n$, is denoted as $T_{n\to n_{\mathrm{diff}}}$. Here, nugget $n_{\mathrm{diff}}$ should be generated in a way that it would still make the turn natural; since it is probable that some dialogue acts would not make sense in the turn. In this context, natural means that the turn should be grammatically correct and keeps the turn understandable to the user. For example, it could be natural to replace a \textbf{Yes Answer} nugget ``Yes, you are correct'' with a \textbf{Agreement} nugget ``I agree'', but it would not be natural to replace this with a \textbf{Disagreement} nugget ``I disagree''. For a nugget $n(\in T)$ in the turn, one such $n_{\mathrm{diff}}$ is generated for every dialogue act that would act naturally in the turn. However, if we take into account the whole generated set of such nuggets $\{n_{\mathrm{diff}}\}$, it is possible that this eliminates the effects of outliers. Therefore, to maintain the effects of outliers, we compute the score $s(T_{n\rightarrow n_{\mathrm{diff}}})$ for each $n_{\mathrm{diff}}$, and take the top $K$ scores; we denote the set of these scores by $S_{\mathrm{diff}}(T,n)$ (where $|S_{\mathrm{diff}}(T,n)|=K)$. Then, the difference between the original turn-level score $s(T)$ and each $s(T_{n\to n_{\mathrm{diff}}})(\in S_{\mathrm{diff}}(T,n))$, is calculated. Then, we represent the mean difference of these scores, $\textit{MD}_{\mathrm{diff}}(T,n)$, where:

\begin{equation}
    \textit{MD}_{\mathrm{diff}}(T,n) = \frac{1}{K} \sum_{s \in S_{\mathrm{diff}}(T,n)} (s(T) - s) .
    \label{eq:MSsub}
\end{equation}

To give a measure of higher precision, we also replace nugget $n(\in T)$ with nuggets of the \textit{same} dialogue act, which we represent as $n_{\mathrm{same}}$. This is done to analyze if the nugget is a suitable option within the dialogue act, to see the effect of the nugget structure, and to examine the use of the language of the nugget. For example, an \textbf{Apology} nugget ``I am sorry'' can be modified to ``I apologize.'' We generate a set of such nuggets $\{n_{\mathrm{same}}\}$, which size should be controlled. Then, similarly to how we determined $K$, to maintain the effects of outliers, we compute scores $s(T_{n\rightarrow n_{\mathrm{same}}})$ for each $n_{\mathrm{same}}$, and take the top $L$ scores; we denote the set of these scores by $S_{\mathrm{same}}(T,n)$ (where $|S_{\mathrm{same}}(T,n)|=L)$. Then, the difference between the original turn-level score $s(T)$ and each $s(T_{n\to n_{\mathrm{same}}})(\in S_{\mathrm{same}}(T,n))$, is calculated. We represent the mean difference score, $\textit{MD}_{\mathrm{same}}(T,n)$, where,
\begin{equation}
    \textit{MD}_{\mathrm{same}}(T,n) = \frac{1}{L} \sum_{s \in S_{\mathrm{same}}(T,n)}( s(T) - s ).
    \label{eq:MSmod}
\end{equation}

Finally, here is how the nugget score of $n(\in T)$ is calculated:
\begin{equation}
    \textit{NS}(T,n) = \Phi (w_{\phi} D_{\phi}(T,n) 
    + w_{\mathrm{diff}} \textit{MD}_{\mathrm{diff}}(T,n)
    + w_{\mathrm{same}} \textit{MD}_{\mathrm{same}}(T,n)
    ) .
    \label{eq:Final}
\end{equation}
$w_\phi$, $w_{\mathrm{diff}}$, and $w_{\mathrm{same}}$ are weights assigned to each of the evaluation features $D_{\phi}(T,n)$, $\textit{MD}_{\mathrm{diff}}(T,n)$, and $\textit{MD}_{\mathrm{same}}(T,n)$ respectively, and $\Phi$ is a normalization function to scale the score; in this paper, we let $\Phi$ be a sigmoid function. We consider it natural to define the relationship of the weights to be $w_\phi\geq w_{\mathrm{diff}}\geq w_{\mathrm{same}}$. This is because the deletion of the nugget, $D_{\phi}(T,n)$, is the most demonstrative indication of the effect of the nugget. $\textit{MD}_{\mathrm{diff}}(T,n)$ is only derived to fine-tune the score, so $w_{\mathrm{diff}}$ should never exceed $w_\phi$. $\textit{MD}_{\mathrm{same}}(T,n)$ is calculated mainly to see the effect of the sentence structure and the use of language; therefore, $w_{\mathrm{same}}$ should not exceed $w_\phi$ nor $w_{\mathrm{diff}}$. Also, since it is probable that the longer the turn, the smaller $D_{\phi}(T,n)$, $\textit{MD}_{\mathrm{diff}}(T,n)$ and $\textit{MD}_{\mathrm{same}}(T,n)$ would become, the values assigned to $w_\phi, w_{\mathrm{diff}}, w_{\mathrm{same}}$ should correlate with the turn length. 

\section{Case Study}

In this section, we will show an example of how the nugget-level score, $NS(T,n)$, is derived. Our example will use the EnDex framework \cite{xu-etal-2022-endex}, a framework that can give a turn-level engagingness score by analyzing a system turn. Our code is available on Github\footnote{\url{https://github.com/RikiyaT/Nugget-Level-Evaluation}}.

Engagingness is the quality of being engaging, which is one of the important qualities of an open-domain dialogue system \cite{10.5555/3304652.3304825, 10.1145/3576840.3578319}. The EnDex framework gives a higher score to the turn when it would make the user want to continue the conversation. Our goal in this example is to achieve the engagingness score of each nugget in the turn.

\begin{figure*}[t]
  \begin{center}
  \includegraphics[width=\textwidth]{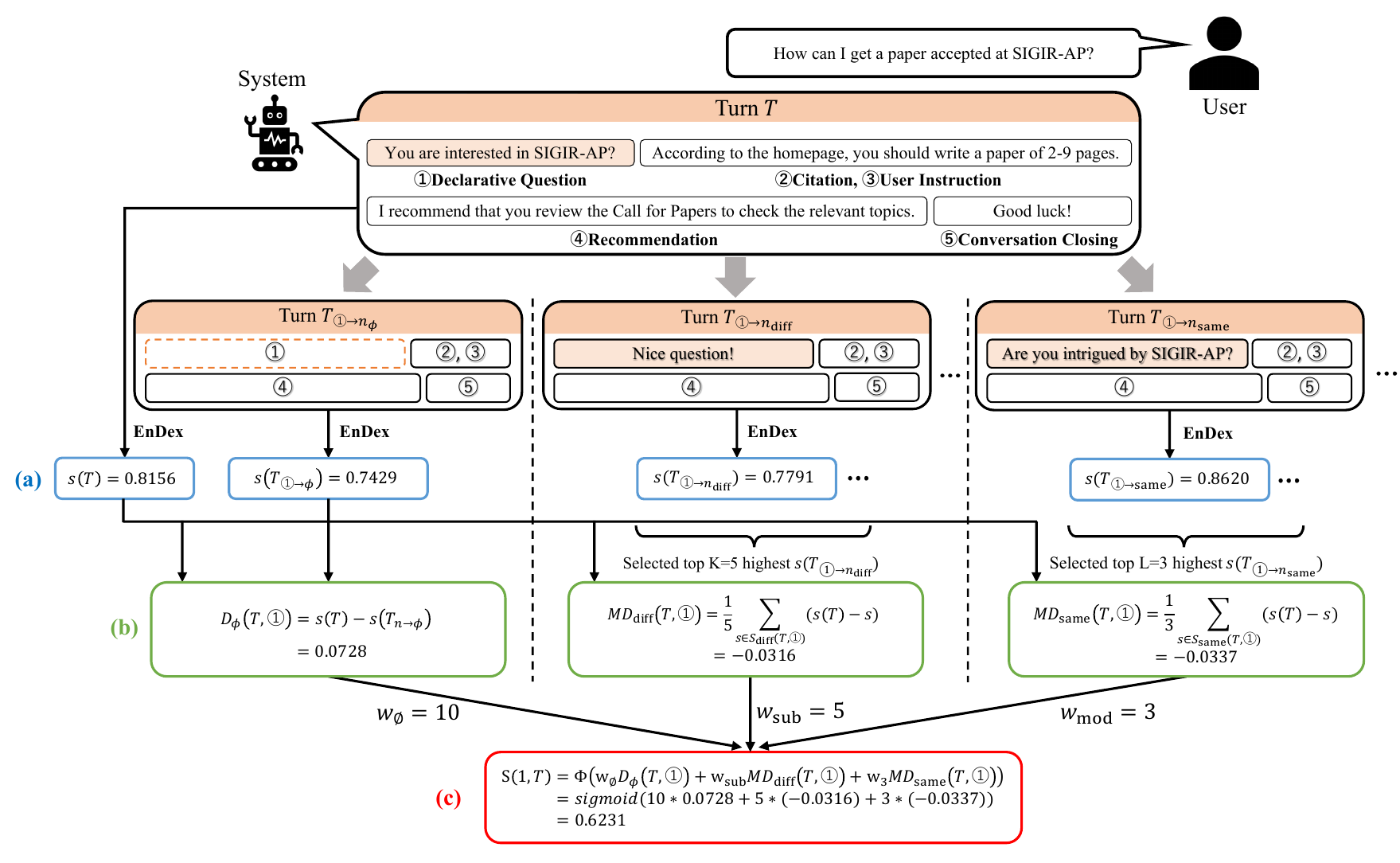}
  \caption{Procedure of Deriving Nugget-Level Score from Turn-Level Score: This figure presents an example of our three-step method to derive the engagingness score of nugget \textcircled{1} from the turn-level engagingness score for a given turn T. (a) depicts the turn-level engagingness scores derived using EnDex framework \cite{xu-etal-2022-endex} of the original turn $T$ and for three versions of the turn, formed by deleting, substituting, and modifying nugget \textcircled{1}. (b) illustrates the calculation of the mean of the score difference value between the original Turn $T$ and each operation-applied turn. Notably, for the nugget-deleted turn, the score is computed by eq. (\ref{eq:MSemp}). In the case of the nugget-substituted and nugget-modified turns, the top $K=5$ values of $s(T_{{\text{\scriptsize\textcircled{1}}}\to n_{\mathrm{diff}}})$ and $L=3$ values of $s(T_{{\text{\scriptsize\textcircled{1}}}\to n_{\mathrm{same}}})$, respectively, are selected, with their average score differences calculated in relation to the original Turn $T$ as in eq. (\ref{eq:MSsub}) and eq. (\ref{eq:MSmod}). Finally, (c) showcases eq. (\ref{eq:Final}), where the nugget-level engagingness score is obtained by normalizing the weighted sum of the averages through a sigmoid function, resulting in a range between 0 and 1. By similarly applying this to all the other nuggets in the turn, our method enables us to have a precise view of which section is positively or negatively affecting the dialogue quality evaluation.}
  \Description{The system response is classified into dialogue acts of the declarative question, citation, user instruction, recommendation, and conversation closing}
  \label{f:fig_new}
  \end{center}
\end{figure*}

In the example, we will analyze a single interaction between a user and a system. We manually classified the system turn into nuggets with corresponding dialogue acts as in Figure~\ref{f:fig_new}. The circled numbers denote the nugget.

For each labeled nugget, we have manually generated nuggets that can be replaced. A nugget $n_{\mathrm{diff}}$ should be a nugget of another dialogue act and should act naturally when placed in the turn. If a nugget $n_{\mathrm{diff}}$ of a certain dialogue act is created, no other nuggets of the same dialogue act can be created. We will show all $n_{\mathrm{diff}}$ of the nugget \textcircled{1}, ``You are interested in SIGIR AP?'', in Table~\ref{tab:n_alt}.

Nuggets $n_{\mathrm{same}}$ are modified nuggets that share the same dialogue acts as the original nugget $n$. In this example, we chose to create seven of these for each nugget $n$. These nuggets are also created in a way they can act naturally when placed in the turn and have the same dialogue act. We will show all $n_{\mathrm{same}}$ for nugget \textcircled{1} of the example turn in Table~\ref{tab:n_reph}.

\begin{table}[t]
    \centering
    \caption{List of $n_{\mathrm{diff}}$ generated to replace nugget \textcircled{1} of the example turn.}
    \begin{tabular}{ccl}
        \toprule
        Dialogue Act & Nugget $n_{\mathrm{diff}}$\\
        \midrule
        Opening & Hello, how are you?\\
        Apology & I apologize for the lack of information.\\
        Rejection & I cannot provide a very detailed answer.\\
        Assumption & I assume you want are interested in IR.\\
        Acknowledgment & Alright, I will answer this for you.\\
        Clarification & You want me to tell you about SIGIR-AP.\\
        User Instruction & Here is what you should do.\\
        Recommendation & I suggest you do the following.\\
        \bottomrule
        \label{tab:n_alt}
    \end{tabular}
\end{table}

\begin{table}[t]
    \caption{List of $n_{\mathrm{same}}$ generated to replace the nugget \textcircled{1} of the example turn. All are `Declarative Question' dialogue acts. We will have seven $n_{\mathrm{same}}$.}
    \begin{tabular}{c}
        \toprule
        Nugget $n_{\mathrm{same}}$\\
        \midrule
        Are you intrigued by SIGIR-AP?\\
        Are you interested in submitting a research paper to SIGIR AP?\\
        Do you want to submit a paper to this conference?\\
        Do you want to know the requirements of SIGIR AP?\\
        Do you want to know about SIGIR AP?\\
        Is your research related to information retrieval?\\
        Are you thinking of attending SIGIR-AP?\\
        \bottomrule
    \end{tabular}
    \label{tab:n_reph}
\end{table}

In this example, we calculate the $NS(T, n)$ score, when $K=5$, $L=3$, and $\{w_\phi, w_{\mathrm{diff}}, w_{\mathrm{same}}\}=\{10, 5, 2\}$. We have chosen these values because they seemed to have a demonstrative result when analyzing this turn on the EnDex framework. 

Figure 1 demonstrates the process of how the nugget-level engagingness score of the first nugget, ``You are interested in SIGIR-AP?'', is calculated using the EnDex framework. First, we delete the first nugget, or we replace it with nuggets $n_{\mathrm{diff}}$ and $n_{\mathrm{same}}$. In step (a), the turn-level engagingness score for the original turn $T$ and the three types of the turn, formed by deleting, substituting, and modifying nugget \textcircled{1} is computed using the EnDex framework. In section (b), we illustrate the computation of the mean score difference between the original Turn $T$ and each of the generated turns. For the nugget-deleted turn, this is calculated as in eq. \ref{eq:MSemp}. In the cases of nugget-substituted and nugget-modified turns, we first select the top $K=5$ values of $s(T_{{\text{\scriptsize\textcircled{1}}}\to n_{\mathrm{diff}}})$ and the top $L=3$ values of $s(T_{{\text{\scriptsize\textcircled{1}}}\to n_{\mathrm{same}}})$, respectively. We then calculate the mean difference between these values and the score of the original Turn $T$. Lastly, section (c) demonstrates the computation of the nugget-level engagingness score. This is achieved by normalizing the weighted linear sum of the average scores for each operation-applied turn, using the sigmoid function to constrain the final nugget-level score between 0 and 1. This method allows for a more granular and normalized evaluation of engagingness at the nugget level, which is particularly useful for open-domain dialogue quality assessment.

Table~\ref{tab:nugget_weight} shows the $NS(T, n)$ score for all five nuggets of the example turn in Figure~\ref{f:fig_new}. From the table, we could notice that the \textbf{Recommendation} nugget, ``I recommend that you review the Call for Papers to check the relevant topics'', is the most engaging nugget. We could analyze that this nugget is engaging because it makes a detailed recommendation of what the user should do. We could also notice the \textbf{User Instruction} nugget, ``You should write a paper of 2-9 pages'', and \textbf{Conversation-Closing}, ``Good Luck!'' are the least engaging nuggets. The \textbf{User Instruction} nugget could seem a little like an order which humans have to obey, which is not engaging. The \textbf{Conversation-Closing} nugget is clearly not engaging because it is trying to end the conversation. 

\begin{table}[t]
\centering
    \caption{Nugget level score $NS(T, n)$ score of the five nuggets in the turn in Figure~\ref{f:fig_new}: The shown scores are nugget-level engagingness scores, derived from the turn-level engagingness scores. The parameters are set to $K=5$, $L=3$, $\{w_\phi, w_{\mathrm{diff}}, w_{\mathrm{same}}\}=\{10, 5, 2\}$. A higher score is more engaging. The circled numbers denote each nugget in the example turn.}
    \label{tab:nugget_weight}
    \begin{tabular}{ccc}
    \hline
    Nugget & $NS(T, n)$ \\
    \hline
        \textcircled{1} & 0.6231 \\
        \textcircled{2} & 0.6294 \\
        \textcircled{3} & 0.3229 \\
        \textcircled{4} & 0.7599 \\
        \textcircled{5} & 0.3892 \\
    \hline
    \end{tabular}
\end{table}

Our case study shows the ability of our method in giving more precise insights by identifying the strengths and weaknesses of the system turn.

\section{Conclusion and Future Work}
In this paper, we focus on evaluating open-domain dialogue systems at the nugget level by using a turn-level evaluation framework. To achieve this, our approach analyzes the contributions of each nugget to the turn-level evaluation score, by calculating the differences in turn-level scores between the original turn and the modified turns (see Figure~\ref{f:fig_new}). In the case study, we show an example application and derive nugget-level engagingness scores. This provides more precise insights by identifying engaging sections and non-engaging sections within each turn. We also introduce new taxonomies for classifying sentences of open-domain dialogue system responses.

Our method is designed to be applicable to various regression-based turn-level evaluation frameworks and multiple dialogue qualities. Therefore, in future work, we plan to explore and validate its effectiveness across these diverse applications. Moreover, it is insightful to investigate the extent to which the performance of our method is influenced by the quality of the turn-level framework employed. Another ambition is to work towards the automation of nugget-level evaluation; this requires the development of automated classification systems and mechanisms for generating adaptable nuggets. Additionally, there is the exciting possibility of applying this evaluation method to a nugget-based open-domain dialogue system evaluation that considers multiple dialogues.

\bibliographystyle{ACM-Reference-Format}
\bibliography{sample-base}

\end{document}